\documentclass{esannV2}
\usepackage[pdftex]{graphicx}
\usepackage[latin1]{inputenc}
\usepackage{booktabs}
\usepackage{amssymb,amsmath,array}
\begin{document}

%style file for ESANN manuscripts
\title{Continual Learning for Human State Monitoring}

%***********************************************************************
% AUTHORS INFORMATION AREA
%***********************************************************************
\author{Federico Matteoni$^1$, Andrea Cossu$^{1,2}$, Claudio Gallicchio$^1$,\\Vincenzo Lomonaco$^1$ and Davide Bacciu$^1$
%
% Optional short acknowledgment: remove next line if non-needed
\thanks{This work has been partially supported by the European Community H2020 programme under project TEACHING (Grant No. 871385).}
%
% DO NOT MODIFY THE FOLLOWING '\vspace' ARGUMENT
\vspace{.3cm}\\
%
% Addresses and institutions (remove "1- " in case of a single institution)
1- University of Pisa - Computer Science Department \\
Largo B. Pontecorvo, 3 - 56127 Pisa - Italy
%
% Remove the next three lines in case of a single institution
\vspace{.1cm}\\
2- Scuola Normale Superiore \\
Piazza dei Cavalieri, 7 - 56126 Pisa - Italy\\
}
%***********************************************************************
% END OF AUTHORS INFORMATION AREA
%***********************************************************************

\maketitle

\begin{abstract}
Continual Learning (CL) on time series data represents a promising but under-studied avenue for real-world applications. We propose two new CL benchmarks for Human State Monitoring. We carefully designed the benchmarks to mirror real-world environments in which new subjects are continuously added. We conducted an empirical evaluation to assess the ability of popular CL strategies to mitigate forgetting in our benchmarks. Our results show that, possibly due to the domain-incremental properties of our benchmarks, forgetting can be easily tackled even with a simple finetuning and that existing strategies struggle in accumulating knowledge over a fixed, held-out, test subject. 
\end{abstract}

\section{Introduction}
Continual Learning (CL) refers to the setting where the data is modeled as a non-stationary stream composed of $n$ experiences $e_0,\ldots,e_n$ \cite{lomonaco2021a}. Each experience is a set of one or multiple samples which are used to perform the training of the model. A CL algorithm processes each experience sequentially and uses them to update the model. One of the major obstacles in learning continuously is the problem of catastrophic forgetting \cite{DBLP:journals/corr/abs-1802-07569}, which induces a lower performance on previous data once the model has been trained on new samples. \\
While catastrophic forgetting is heavily studied in contexts like computer vision \cite{DBLP:journals/corr/abs-2109-11369}, its impact on alternative environments remains under-documented. In particular, time-series data and recurrent neural networks are fundamental in many non-stationary environments \cite{COSSU2021607, cossu2021} and a deeper understanding of the behavior of these models may facilitate the development of novel CL applications. \\
One of the major limitations in the study of CL problems with time series data is the lack of standard benchmarks and datasets against which to evaluate the performance of existing and novel strategies.\\
To address this problem, in this paper we focus on Human Activity Recognition tasks \cite{JHA20211} from time-series data. We introduced two Human State Monitoring benchmarks for CL. Our benchmarks fall into the domain-incremental scenario \cite{vandeven2018}, where each experience provides new data for already seen classes (the different human states). We ran an extensive empirical evaluation on our benchmarks to assess 1) the impact of catastrophic forgetting with respect to different CL strategies and 2) the ability of existing strategies to accumulate knowledge over time and improve their performance as new data is encountered.
%Our results show that WESAD is effectively tackled by recurrent models and that adding replay is beneficial for knowledge accumulation on a fixed test set. Even if ASCERTAIN represents a more challenging benchmark for RNNs, where they achieve lower accuracy, forgetting is present but not catastrophic across all our experiments. The domain-incremental nature of our benchmarks allows to shift the focus from the mitigation of forgetting, which constitutes the vast majority of the current CL literature, to additional metrics like knowledge accumulation and forward transfer.

\begin{table}[t]
  \centering
  \small
  \begin{tabular}{l c c c}
    \toprule
         & \textbf{WESAD} & \textbf{ASCERTAIN} & \textbf{Cust. ASCERTAIN}\\ \midrule
        \textbf{Subjects} & 15 & 17 & 17 \\
        \textbf{Classes} & 4 & 4 & 4 \\
        \textbf{Sequence length} & 100 & 160 & 160 \\
        \textbf{Features} & 14 & 18 & 18 \\
        \textbf{Training set size} & 4500 & 2160 & 1836 \\  % custom da verificare
        \textbf{Test set size} & 1500 & 720 & 720 \\ % custom da verificare
    \bottomrule
    \end{tabular}
  \caption{Datasets summary.}\label{Tab:Datasets}
\end{table}

\section{Two Benchmarks for Continual Human State Monitoring} \label{sec:benchmark}
We propose two new benchmarks for the evaluation of CL approaches on time series data. In particular, we focus on the Human State Monitoring task, which provides a natural source of non-stationary, sequential activities in which time plays a fundamental role. 
Our benchmarks are based on two existing datasets for time-series classification: WESAD \cite{10.1145/3242969.3242985} and ASCERTAIN \cite{7736040}. Both datasets provide sequences of physiological data and each sequence is classified in 4 labels that denote the self-evaluated mental state of each subject. WESAD provides temperature and movement data, in addition to electromiogram (EMG) data and electrodermal (EDA) data, while ASCERTAIN provides electroencephalogram (EEG) and galvanic skin response (GSR) data. Both datasets provide heartbeat data via ECG measurements.
Crucially for the design of our benchmarks, WESAD and ASCERTAIN are organized by subject. We used this property to build a non-stationary environment for CL: for both datasets we created a stream in which each experience brings data coming from two subjects. The CL models we tested are trained sequentially on each experience of the stream. For each benchmark, we held out one subject to be used as test set. The task is to classify each time-series into one of the possible classes, which does not vary across experiences. 
We applied a preprocessing pipeline to WESAD by resampling each sequence at 32Hz and by normalizing mean and variance of each feature. Following WESAD documentation, we removed the labels 0 (neutral), 5, 6, 7 and regrouped each sequence into subsequences of 100 points per label. For each label, we kept 100 subsequences, leaving us with a training set of 4500 elements and a test set of 1500 elements.
For ASCERTAIN, we ignored subjects 44 and 52 whose data quality is highlighted in the dataset as poor. For each remaining subject, we created the labels based on valence and arousal self-reported levels. Similar to WESAD, we resampled each sequence at 32Hz, we removed subjects with missing features and we built subsequences of 160 points. We ended up with a training set of 2160 elements and a test set of 720 elements.
The ASCERTAIN dataset contains an over representation of the class 0: while the classes 1, 2 and 3 are represented each in around 500 sequences, the class 0 is represented in around 1100 sequences. To assess the impact of unbalanced classes in ASCERTAIN, we produced a custom version by artificially balancing it, removing sequences from over represented class. The preprocessing follows the one used for original ASCERTAIN. Table \ref{Tab:Datasets} summarizes the main characteristics of the three datasets.

\begin{table}[t]
  \centering
  \small
  \begin{tabular}{lccc}
    \toprule
        \textbf{Strategy} & \textbf{WESAD} & \textbf{ASCERTAIN} & \textbf{Cust. ASCERTAIN}\\ \midrule
        \textbf{Offline} & 99.60\%$_{\pm 0.14\%}$ & 30.39\%$_{\pm 1.74\%}$ & 39.09\%$_{\pm 0.13\%}$ \\ \midrule\midrule
        \textbf{Naive} & 72.41\%$_{\pm 4.50\%}$ & 26.78\%$_{\pm 2.47\%}$ & 32.77\%$_{\pm 7.01\%}$ \\ \midrule
        \textbf{Cumulative} & 83.62\%$_{\pm 10.47\%}$ & 29.59\%$_{\pm 2.38\%}$ & 35.62\%$_{\pm 4.51\%}$ \\ \midrule
        \textbf{Replay} & 80.79\%$_{\pm 7.95\%}$ & 26.75\%$_{\pm 2.18\%}$ & 33.07\%$_{\pm 4.59\%}$ \\ \midrule
        \textbf{EWC} & 71.79\%$_{\pm 4.74\%}$ & 26.66\%$_{\pm 2.17\%}$ & 34.46\%$_{\pm 6.12\%}$ \\ \midrule
        \textbf{LwF} & 72.75\%$_{\pm 3.82\%}$ & 27.62\%$_{\pm 2.36\%}$ & 33.03\%$_{\pm 6.12\%}$ \\ \bottomrule
    \end{tabular}
  \caption{Final average accuracy and standard deviation per dataset over 5 runs.}\label{Tab:FinalAccuracy}
\end{table}

\section{Empirical Evaluation}
\begin{figure}
    \centering
    \includegraphics[scale=0.29]{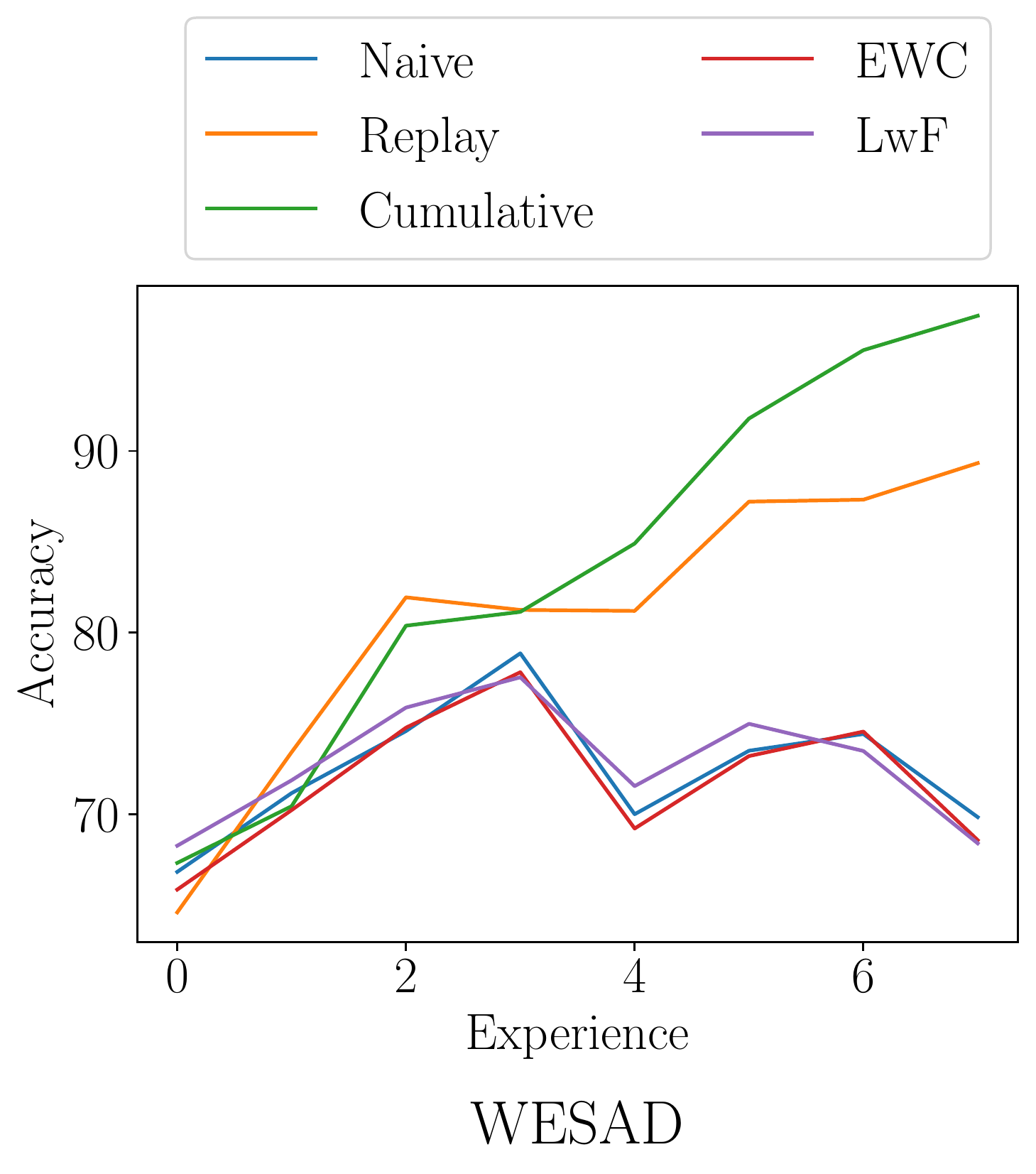}
    \hfill
    \includegraphics[scale=0.24]{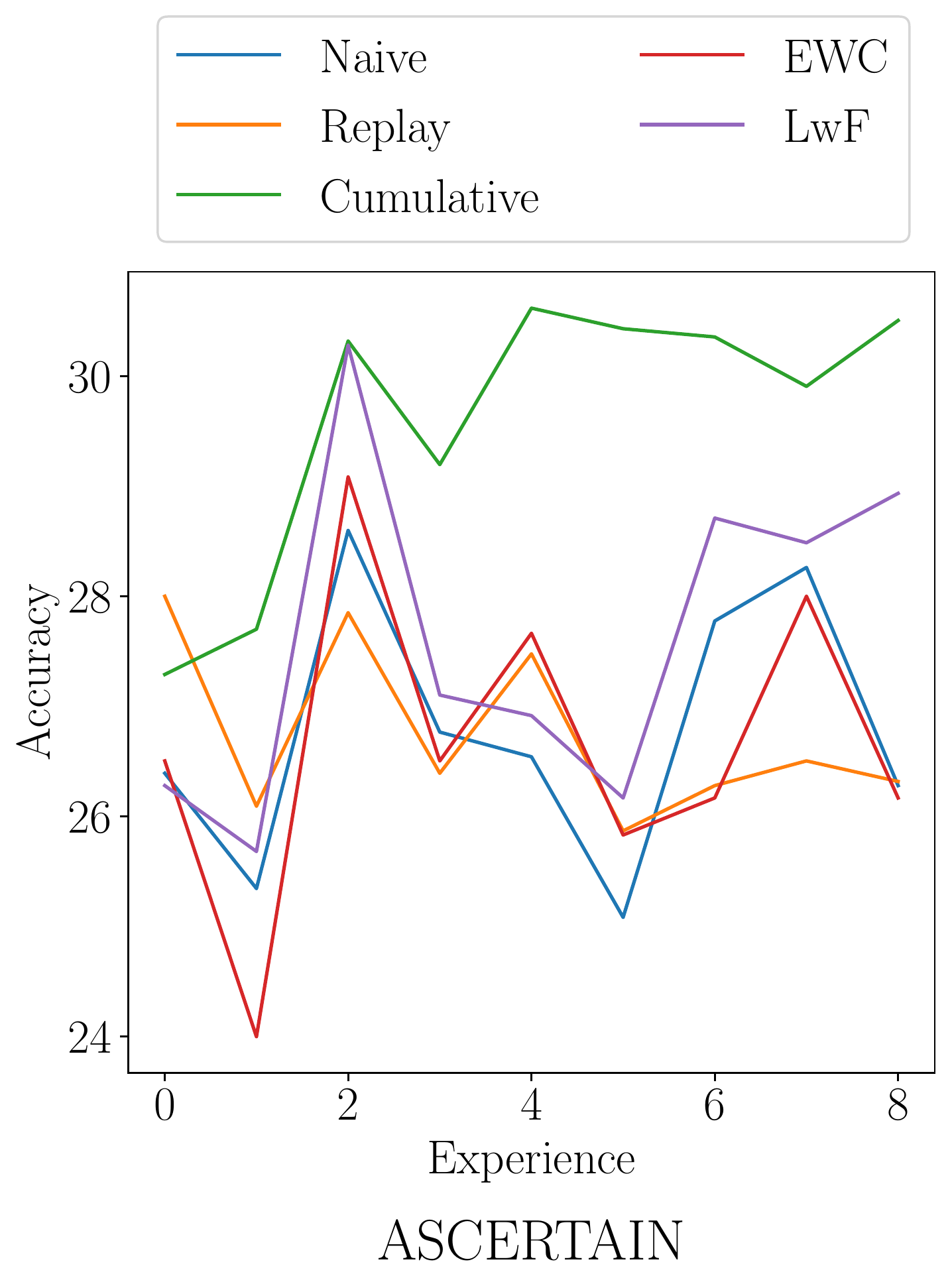}
    \hfill
    \includegraphics[scale=0.24]{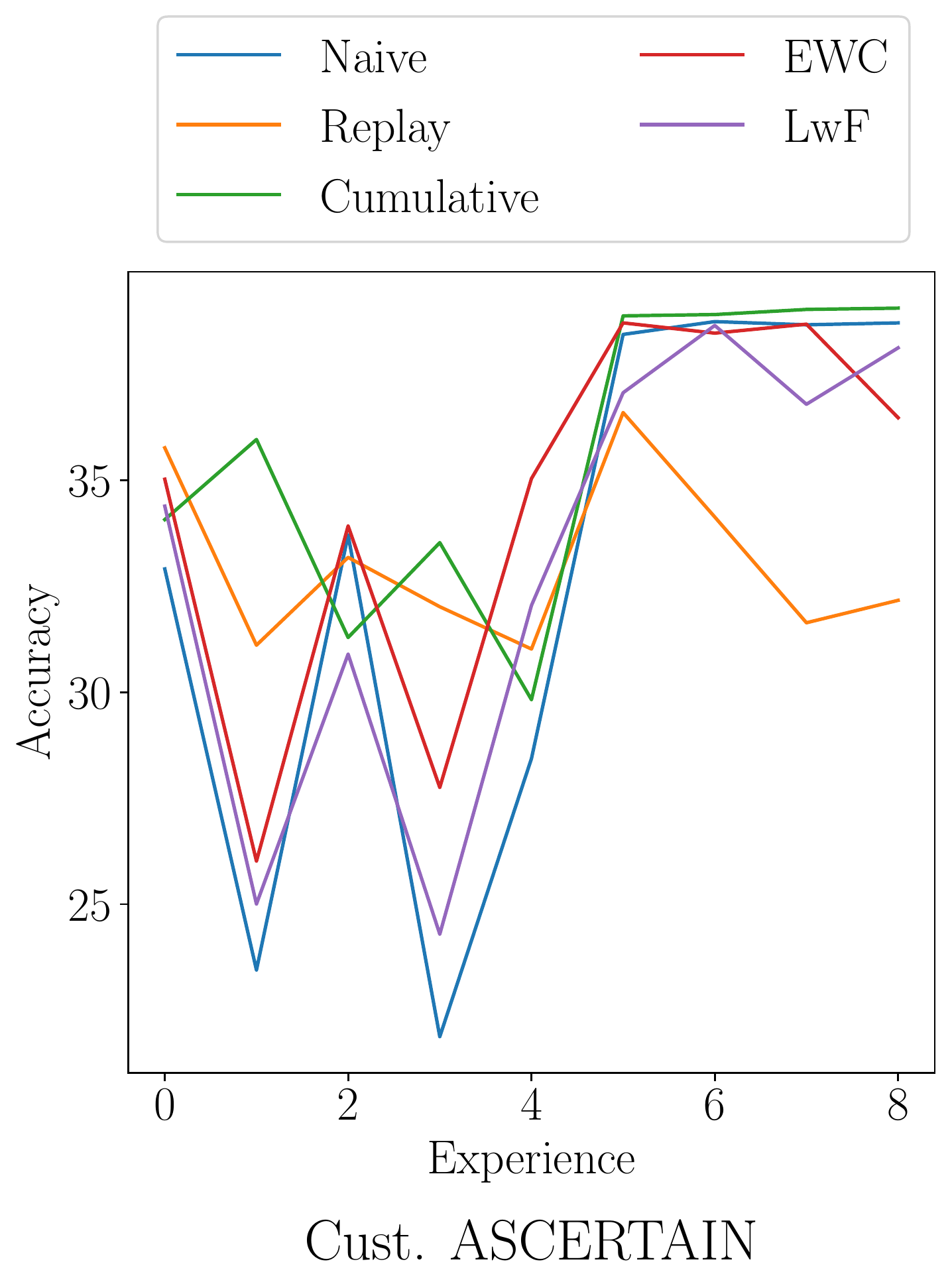}
    \caption{Accuracy on each dataset measured on the held-out test set after training on each experience}
    \label{fig:acc}
\end{figure}

We conducted an empirical evaluation to assess the behavior of different CL strategies applied to the proposed benchmarks. Our main objective is to understand to what extent training continuously on novel subjects affects the performance on the held-out test subject. That is, we aim to understand if and how much recurrent neural networks suffer from catastrophic forgetting in our proposed benchmarks.
In order to enable full reproducibility of our results, here we briefly describe our experimental setup, and we also publicly release the code\footnote{https://github.com/fexed/CLforHSM} used in all the experiments, which are conducted through the Avalanche library\cite{lomonaco2021a}.
\paragraph{Experiments setup} We tested four popular CL strategies: Replay \cite{hayes2021a} keeps a percentage of the previous training set to be used in the following experience, Elastic Weight Consolidation (EWC) \cite{doi:10.1073/pnas.1611835114} limits the change in model parameters based on an importance value associated to each parameter, Learning without Forgetting (LwF) \cite{DBLP:journals/corr/LiH16e} uses Knowledge Distillation to stabilize the model activations with respect to the model at the previous experience. We also provide results for the Naive and Cumulative strategies. Naive acts as a lower bound since it finetunes the model across the experiences without any CL technique. Cumulative is a special case of replay in which all previous data is used at each experience. Finally, the offline (joint training) approach acts as upper bound for the CL performance since it consists of a single training phase in which the union of data in all the experiences is used. As such, offline training does not constitute a valid CL strategy.
We used a Recurrent Neural Network composed by 2 layers of 18 GRU units for both datasets. We conducted model selection on the offline training and the best resulting model has been used to train all the CL strategies used in our empirical evaluation: the hyperparameters involved in the model selection were the learning rate, the $\beta_1,\beta_2$ of the Adam optimizer and the L1L2 regularizer hyperparameter. The replay memory size has been set at 25\% of the training set size, which is 70 patterns for WESAD and 25 patterns for ASCERTAIN. For each experiment, we monitored the average accuracy and average epoch training time and computed mean and standard deviation over 5 runs.
\begin{figure}
    \centering
    \includegraphics[width=\textwidth]{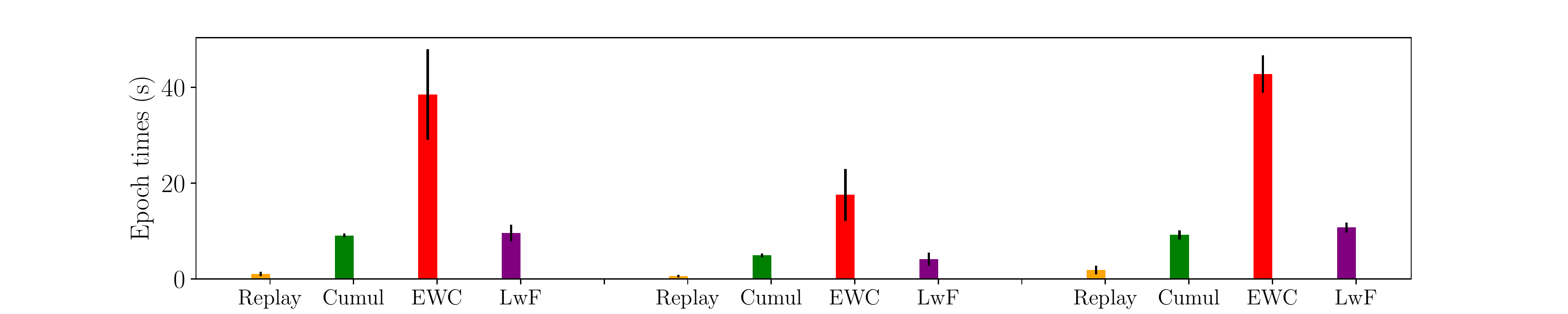}
    \caption{Epoch times of CL strategies on each dataset (in seconds). Naive not reported since it does not use any CL technique. It is the least expensive across all experiments.}
    \label{fig:alltimes}
\end{figure}
\paragraph{Results} Table \ref{Tab:FinalAccuracy} reports the average accuracy on the test set at the end of training on the entire stream of experiences, while Figure \ref{fig:acc} shows the average accuracy for each benchmark after training on each experience. We also reported the average training epoch time in Figure \ref{fig:alltimes} to properly compare the computational cost of each strategy. The WESAD benchmark is effectively tackled by RNNs. Both Cumulative and Replay strategies are able to accumulate knowledge over time, as the model is trained on more experiences. The gap with offline performance is still present, although it is smaller than the one achieved by the other strategies. There is no clear advantage in the use of regularization strategies like LwF and EWC with respect to a Naive finetuning approach. While this may be surprising in a traditional class-incremental setting, in our domain-incremental benchmark \cite{vandeven2018} the deterioration of performance due to catastrophic forgetting is much less severe. 
Both ASCERTAIN and its custom, balanced version represent a more difficult task with respect to WESAD. The offline performance is low, highlighting the fact that RNNs struggle to properly learn this kind of task. Balancing the dataset, as in Custom ASCERTAIN, improves the offline accuracy of around 9 points.
The CL strategies still show a gap with respect to offline training. However, in the ASCERTAIN cases, the application of Replay does not contribute to improve the final performance compared to the other strategies. This is strongly related to the low accuracy achieved during offline training.

\paragraph{Discussion} Our experiments show that recurrent models do not suffer from catastrophic forgetting in any of our benchmarks for human state monitoring. This is mainly due to the domain-incremental CL scenario defined by the benchmarks. Especially in the case of time-series, it is likely that the stream of incoming data will not always introduce new classes, but rather new instances of the same task that was required to solve from the beginning. Our benchmarks fit this case and will contribute to model more real-world applications where, rather than forgetting, the focus is mostly on knowledge accumulation and forward transfer. As showed by Figure \ref{fig:acc}, regularization strategies like LwF and EWC are not capable of accumulating knowledge, since they are mostly designed to improve model stability against forgetting. To this end, replay is the most effective strategy with respect to the alternatives, also in terms of time complexity during training (Figure \ref{fig:alltimes}). However, storing previous data may not be possible in all applications due to data-privacy concerns or other constraints. This highlights the need to develop novel CL strategies which are capable of rapidly exploiting new information to improve on the current task. 

\section{Conclusion and Future Works}
We proposed two new CL benchmarks for Human State Monitoring on time-series data. Both benchmarks introduce new subjects sequentially and keep one held-out subject for test, a common case in real-world applications. We used the two benchmarks to produce an empirical assessment of the behavior of common CL strategies. The results show that, on one side, all strategies are able to mitigate forgetting while, on the other side, most of them fail to accumulate knowledge over time (with one exception being replay). The former behavior is due to the domain-incremental nature of the benchmarks, while the latter is due to the fact that most of the existing strategies are designed with the only purpose of mitigating catastrophic forgetting, at the expenses of other important CL objectives like forward transfer. Future works may extend our benchmarks with a larger number of subjects, in order to better study knowledge accumulation over longer streams of data. Our work also highlights the need to design CL strategies which are able to exploit new data to improve their performance on unseen samples, rather than only focusing on forgetting. We believe Human State Monitoring to be a promising application for real-world CL with time-series data and that its study will open new challenges and opportunities for continual learning agents.

\begin{footnotesize}

\bibliographystyle{unsrt}
\bibliography{Bibliografia}

\end{footnotesize}

\end{document}